\documentclass[runningheads]{llncs}

\usepackage{accv}

\usepackage{accvabbrv}

\usepackage{graphicx}
\usepackage{booktabs}

\usepackage[accsupp]{axessibility}  

\usepackage[pagebackref,breaklinks,colorlinks,citecolor=accvblue]{hyperref}

\usepackage{orcidlink}

\begin{document}

\title{SwinIFS: Landmark‑Guided Swin Transformer For Identity‑Preserving Face Super‑Resolution} 


\author{Habiba Kauser\inst{1}\and
Saeed Anwar\inst{2}\and
Omar Hammad\inst{1}\and \\
Ibrahim Radwan\inst{3}\and
Abdul Bais\inst{4}
}

\authorrunning{H. Kausar~et~al.}
\institute{King Fahd University of Petroleum and Minerals, KSA \and
The University of Western Australia, Australia\\
\and
The University of Canberra, Australia\\ \and
The University of Regina, Canada\\
}

\maketitle

\begin{abstract}
  Face super-resolution aims to recover high-quality facial images from severely degraded low-resolution inputs, but remains challenging due to the loss of fine structural details and identity-specific features. This work introduces SwinIFS, a landmark-guided super-resolution framework that integrates structural priors with hierarchical attention mechanisms to achieve identity-preserving reconstruction at both moderate and extreme upscaling factors. The method incorporates dense Gaussian heatmaps of key facial landmarks into the input representation, enabling the network to focus on semantically important facial regions from the earliest stages of processing. A compact Swin Transformer backbone is employed to capture long-range contextual information while preserving local geometry, allowing the model to restore subtle facial textures and maintain global structural consistency. Extensive experiments on the CelebA benchmark demonstrate that SwinIFS achieves superior perceptual quality, sharper reconstructions, and improved identity retention; it consistently produces more photorealistic results and exhibits strong performance even under $8\times$ magnification, where most methods fail to recover meaningful structure. SwinIFS also provides an advantageous balance between reconstruction accuracy and computational efficiency, making it suitable for real-world applications in facial enhancement, surveillance, and digital restoration. Our code, model weights, and results are available at https://github.com/Habiba123-stack/SwinIFS.
\end{abstract}

\section{Introduction}
\label{sec:intro}

Face super-resolution (FSR) aims to reconstruct high-resolution (HR) facial images from low-resolution (LR) inputs while preserving structural coherence and identity-specific details. Reliable recovery of facial features is essential for applications such as surveillance, biometrics, forensics, video conferencing, and media enhancement~\cite{wang2019deep, jiang2021deep}. Unlike generic super-resolution~\cite{anwar2020deep,chen2023swinfsr}, FSR benefits from the strong geometric regularity of human faces, where the spatial arrangement of key components (eyes, nose, mouth) provides valuable prior information for reconstruction. The LR observation process is typically modeled as
\begin{equation}
    I_\mathrm{LR}=\downarrow_{s}(I_\mathrm{HR} \ast k) + \eta,
    \label{eq:1}
\end{equation}

\begin{figure}[t]
\centering
\begin{tabular}{ccccccccc}
\includegraphics[width=0.112\textwidth]{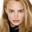}&  
\includegraphics[width=0.112\textwidth]{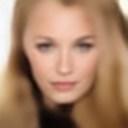}& 
\includegraphics[width=0.112\textwidth]{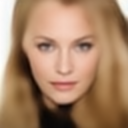}& 
\includegraphics[width=0.112\textwidth]{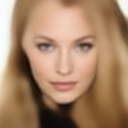}& 
\includegraphics[width=0.112\textwidth]{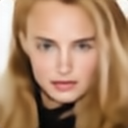}&
\includegraphics[width=0.112\textwidth]{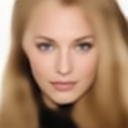}& 
\includegraphics[width=0.112\textwidth]{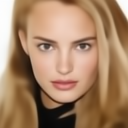}& 
\includegraphics[width=0.112\textwidth]{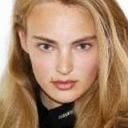}\\
\includegraphics[width=0.112\textwidth]{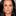}&  
\includegraphics[width=0.112\textwidth]{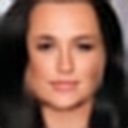}& 
\includegraphics[width=0.112\textwidth]{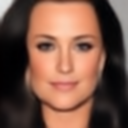}& 
\includegraphics[width=0.112\textwidth]{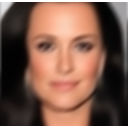}& 
\includegraphics[width=0.112\textwidth]{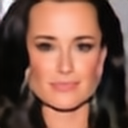}&
\includegraphics[width=0.112\textwidth]{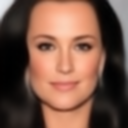}& 
\includegraphics[width=0.112\textwidth]{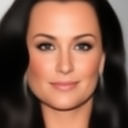}& 
\includegraphics[width=0.112\textwidth]{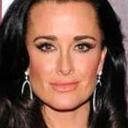}\\
 LR & SRGAN  & DIC & SISN & WIPA & W-NET &Ours & HR \\    
\end{tabular}

\caption{A comparison of face super-resolution results is presented, with the top row illustrating results for a $4\times$ upscaling and the bottom row for an $8\times$ upscaling. Our approach exhibits superior detail recovery and preservation of structure under both degradation conditions.}
\label{fig:x8_visual_sample}
\end{figure}
\noindent where $I_\mathrm{HR}$ is the HR image, $I_\mathrm{LR}$ is the LR image, $k$ denotes the blur kernel, $\downarrow_{s}$ is the downsampling operator, and $\eta$ represents noise. In practical environments, degradation is further compounded by compression artifacts, illumination variations, and sensor noise. At moderate upscaling factors (e.g., $4\times$), some structural cues remain; however, at extreme scales (e.g., $8\times$ inputs), most identity cues are lost, rendering the reconstruction highly ill-posed.

Early face hallucination methods relied on interpolation, example-based patch retrieval, or sparse coding~\cite{baker2000hallucinating}. Although pioneering, these approaches produced overly smooth results and lacked robustness to domain variation. The introduction of deep learning significantly advanced SR performance. CNN-based methods~\cite{dong2014learning, kim2016accurate,lim2017enhanced} improved texture reconstruction but remained limited by their local receptive fields, often leading to globally inconsistent facial structures.

Generative adversarial networks (GANs) improved perceptual realism by learning to synthesize sharper textures~\cite{ledig2017photo}. FSRNet~\cite{chen2018fsrnet} and Super-FAN~\cite{bulat2018superfan} demonstrated that combining GAN objectives with facial priors such as landmarks or parsing maps enhances structural alignment. However, GAN-based methods are susceptible to hallucinating unrealistic details and may compromise identity preservation, especially when LR inputs are highly degraded. Transformer architectures have recently emerged as powerful tools for image restoration due to their ability to capture long-range dependencies through self-attention \cite{dosovitskiy2021an, vaswani2017attention}. The Swin Transformer~\cite{liu2021swin} introduces hierarchical window-based attention, offering an effective balance of global modeling and computational efficiency. Despite their strengths, Transformers alone struggle when key facial cues are absent in severely degraded inputs. Incorporating explicit geometric priors can alleviate this ambiguity.

Facial landmarks provide compact, reliable structural information about the geometry of key facial regions. When encoded as heatmaps, they provide spatial guidance to maintain feature alignment, facial symmetry, and identity consistency during reconstruction~\cite{chen2018fsrnet, yu2018face}. Motivated by these insights, this work proposes a landmark-guided multiscale Swin Transformer framework designed to address both moderate ($4\times$) and extreme ($8\times$) FSR scenarios.

Our proposed method fuses RGB appearance information with landmark heatmaps to jointly model facial texture and geometry. The Swin Transformer backbone captures global contextual relationships, while landmark priors enforce structural coherence. This unified approach enables robust reconstruction across multiple upscaling factors and significantly improves identity fidelity. Experiments on CelebA demonstrate that the proposed framework achieves superior perceptual quality, structural accuracy, and quantitative performance compared to representative CNN, GAN, and Transformer-based baselines.

\section{Related Work}
Face super-resolution has advanced significantly over the past two decades, transitioning from early interpolation schemes to modern deep-learning, adversarial, and transformer-based frameworks. Unlike generic single-image super-resolution (SISR), FSR requires strong preservation of identity and facial geometry, making structural modeling a core research challenge~\cite{Ma_2020_CVPR, chen2018fsrnet}. Early work relied on interpolation and example-based methods~\cite{freeman2002example, park2003super}. Although computationally efficient, these approaches produced overly smooth textures and failed to recover high-frequency facial details. Learning-based extensions, including sparse coding and manifold models~\cite{yang2010image, baker2000hallucinating}, partially improved texture synthesis but struggled under severe degradation and exhibited limited generalization.

Deep learning significantly advanced FSR performance. CNN-based architectures such as SRCNN~\cite{dong2014learning}, VDSR~\cite{kim2016accurate}, and EDSR~\cite{lim2017enhanced} demonstrated that hierarchical feature learning could outperform traditional methods. Face-specific extensions, including FSRNet~\cite{chen2018fsrnet} and URDGN~\cite{chen2018fsrnet, yu2016ultra}, incorporated structural priors, such as landmark heatmaps or facial parsing maps. These models improved alignment and structural consistency, but their reliance on pixel-wise losses often produced smooth outputs and limited high-frequency synthesis.

The introduction of GAN frameworks shifted the focus toward perceptual realism. SRGAN~\cite{ledig2017photo} demonstrated sharper textures using adversarial and perceptual losses. Face-specific GAN models such as Super-FAN~\cite{bulat2018superfan}, FSRGAN~\cite{wang2019fsrgan}, and DICGAN~\cite{bulat2018superfan, Ma_2020_CVPR} incorporated identity losses, alignment modules, or cycle consistency to improve realism and identity preservation. While effective, GAN-based FSR remains sensitive to training instability and may hallucinate unrealistic facial features under extreme downsampling. Recent studies have also explored attention mechanisms and lightweight architectures to enhance reconstruction quality while preserving computational efficiency. SPARNet~\cite{sparnet} incorporated spatial attention mechanisms to improve the recovery of essential facial structures without dependence on explicit facial priors. MRRNet~\cite{huang2022mrrnet} utilized multiscale receptive-field features with spatial attention to enhance local detail reconstruction, thereby avoiding expensive facial prior estimation. Similarly, UFSRNet~\cite{wang2024ufsrnet} introduced a lightweight U-shaped architecture that delivers competitive reconstruction performance with relatively few parameters; nonetheless, its convolution-based design exhibits limited capacity to model long-range contextual dependencies.

Recent advances in face super-resolution have explored richer structural priors and transformer-based architectures to improve reconstruction quality. Wang \textit{et al.}~\cite{wang2023parsing} proposed a facial parsing-guided framework that exploits parsing maps and attention fusion to enhance structural consistency during reconstruction. Jois \textit{et al.}~\cite{jois2024reference} introduced a reference-based face super-resolution approach using a spatial transformer for stable feature alignment with high-resolution reference images. Li \textit{et al.}~\cite{li2025sanet} presented SANet, which integrates self-similarity priors and hybrid attention to capture non-local facial dependencies. Yang \textit{et al.}~\cite{yang2025gvtnet} further improved global facial modeling through a graph vision transformer that explicitly models relationships among facial components. These recent methods demonstrate the growing importance of attention mechanisms, structural priors, and transformer-based representations for face super-resolution, while efficient integration of geometric information for robust identity preservation remains an active research direction.

More recently, attention and transformer-based methods have advanced FSR by modeling long-range dependencies. Vision Transformers~\cite{dosovitskiy2021an} introduced global patch-based attention, but their high computational cost limited their use for low-level restoration. The Swin Transformer~\cite{liu2021swin} addressed this by employing hierarchical shifted-window attention, enabling efficient modeling of global context. Several FSR methods have since incorporated transformer modules, including FaceFormer~\cite{zhang2023faceformer}, and W-Net~\cite{liu2025w}, which combine attention with CNN branches or semantic priors. These approaches achieve strong perceptual and structural performance but often require large memory and long training times, and are typically trained for a single upscaling factor. Moreover, explicit geometric priors, such as facial landmarks, remain underutilized in many transformer-based designs despite their effectiveness at guiding facial structure~\cite{chen2018fsrnet, yu2018face}.

Overall, existing CNN and GAN methods struggle to balance high-frequency detail reconstruction with identity fidelity. At the same time, transformer models provide superior global modeling, but at the cost of increased complexity and limited structural conditioning. These limitations motivate a unified approach that integrates explicit landmark priors with an efficient Swin Transformer backbone to improve structural coherence, identity preservation, and multi-scale robustness in face super-resolution. In addition, recent studies emphasize the growing need for multi-scale FSR systems capable of handling diverse real-world degradations such as compression, occlusion, and significant pose variation. Most current models are trained on a single fixed scale or under controlled laboratory conditions, limiting their generalization to practical scenarios in which facial resolution varies widely. Moreover, despite their demonstrated value in CNN and GAN architectures, structural priors are rarely embedded deeply into transformer backbones. This gap highlights an opportunity for new frameworks that seamlessly integrate geometric cues with global attention to achieve stable, identity-consistent reconstruction across moderate and extreme upscaling factors.

\section{SwinIFS}
Face super-resolution is an inherently ill-posed problem, as a single low-resolution input may correspond to multiple plausible high-resolution facial configurations. This ambiguity arises because the LR image lacks fine-grained texture details, subtle identity cues, and structural regularities present in HR images. To resolve this, our methodology integrates structural priors from facial landmarks with the hierarchical modeling capabilities of Swin Transformers. Landmark heatmaps provide explicit geometric guidance, ensuring that the network focuses on identity-sensitive regions such as the eyes, nose, and mouth. Simultaneously, Swin Transformers enable global spatial reasoning by capturing both localized texture patterns and long-range dependencies across facial regions.

\begin{figure}[t]
\centering 
\includegraphics[width=\textwidth]{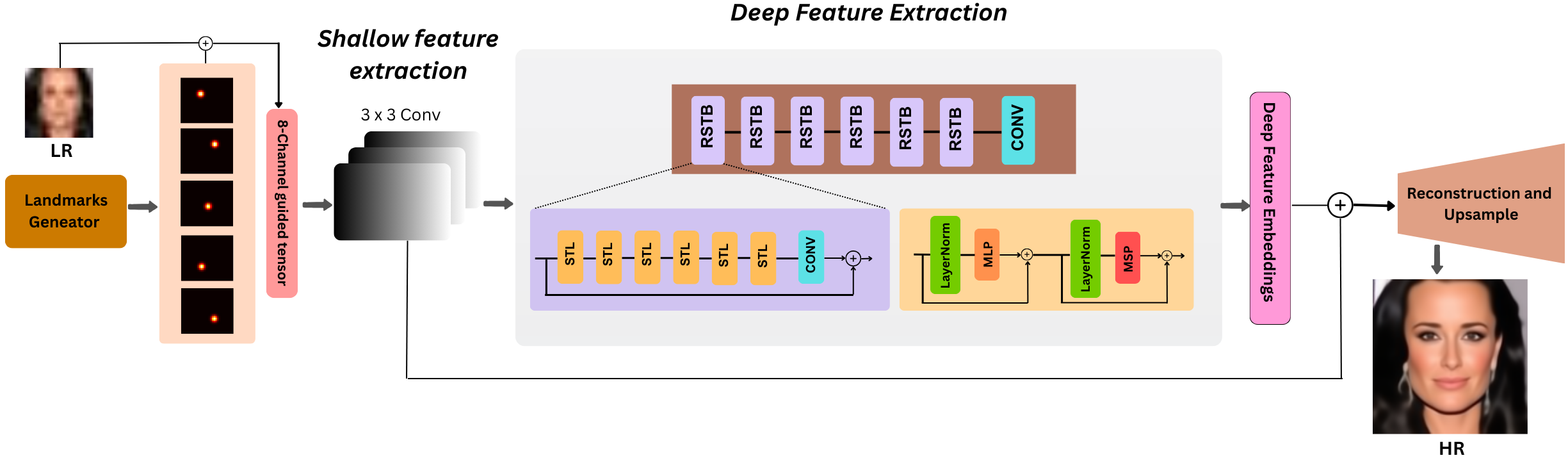} 
\caption{ Overview of the proposed SwinIFS framework. The LR face image is combined with five Gaussian landmark heatmaps to form an 8-channel input tensor. The features are refined through stacked Swin Transformer layers and RSTBs, followed by PixelShuffle reconstruction to generate the final high-resolution face image. } 
\label{fig:method_SwinIFS} 
\end{figure}

The overall pipeline is illustrated in Figure~\ref{fig:method_SwinIFS}. The framework consists of four stages: landmark-guided input construction, feature extraction, transformer-based refinement using Residual Swin Transformer Blocks (RSTBs), and high-resolution reconstruction via PixelShuffle upsampling.

\subsection{Landmark Encoding and Input Construction}
To integrate meaningful geometric information into the network, we begin by generating a low-resolution image from an aligned high-resolution input using bicubic interpolation, $I_{\mathrm{LR}} = \downarrow_{S}(I_{\mathrm{HR}}),~where~S \in \{4,8\}$. This provides the baseline visual input. However, LR faces often lack crucial structural cues, making it difficult for SR models to infer identity-consistent high-frequency content. To mitigate this, we extract five key landmarks (left eye, right eye, nose, and mouth corners) and convert each point into a Gaussian heatmap,
$M_i$. These heatmaps produce a soft, spatially aware representation that indicates the locations of critical facial components, rather than providing only discrete landmark coordinates. Stacking the five heatmaps yields $M_{\mathrm{c}} \in \mathbb{R}^{C\times H \times W}$, where $C=5$. Finally, the LR RGB image and the landmark maps are concatenated:
\begin{equation}
I_{\mathrm{in}} = [I_{\mathrm{LR}}|| M_{\mathrm{c}}],
\label{eq:2}
\end{equation}
where $||$ stands for concatenation. This produces an 8-channel tensor that explicitly encodes both appearance and geometry, allowing the network to fuse structural priors and texture information from the earliest stages of processing. By embedding geometry directly into the input, the model avoids depending solely on visual cues that may be missing or ambiguous in LR images.

\subsection{Shallow and Deep Feature Extraction}
The SwinIFS network begins processing this 8-channel input by projecting it into a high-dimensional feature space using a convolution $H_{\mathrm{SF}}$:

\begin{equation}
F_0 = H_{\mathrm{SF}}(I_{\mathrm{in}}),
\label{eq:3}
\end{equation}
where $F_0$ is the extracted features. This preserves spatial resolution while expanding representational capacity. The shallow features capture local edges, coarse textures, and the spatial distribution of the landmark heatmaps. These encoded cues serve as the foundation for deeper reasoning. Next, the feature tensor is passed through a hierarchy of $D$ stacked Residual Swin Transformer Blocks (RSTBs). Each RSTB learns progressively more complex semantic information, building from local texture patterns in early layers to global structure and identity-relevant features in deeper layers. The pipeline follows a recursive formulation:
\begin{equation}
F_i = \mathrm{RSTB}_i(F_{i-1}),
\label{eq:4}
\end{equation}
within each block, multiple Swin Transformer Layers (STLs) refine the feature maps:
\begin{equation}
F_{i,j} = \mathrm{STL}_{i,j}(F_{i,j-1}).
\label{eq:5}
\end{equation}

Swin Transformer Layers divide the feature map into local windows and compute multi-head self-attention within each window. This operation enables the model to selectively enhance relevant regions based on their spatial and contextual relationships. Alternating between regular and shifted window partitions allows cross-window communication, effectively expanding the receptive field. Thus, the model learns global facial geometry (overall head shape and symmetry) and fine structural relationships (eye distance and mouth curvature) simultaneously. To preserve stability and prevent loss of low-frequency content, a global skip connection merges shallow and deep features:
\begin{equation}
F_{\mathrm{res}} = F_0 + H_{\mathrm{Conv}}(F_D).
\label{eq:6}
\end{equation}

This fusion ensures that early structural cues from the input remain intact while deeper layers refine high-frequency textures and identity-specific details.

\subsection{Residual Swin Transformer Block}

The RSTB is the fundamental module enabling SwinIFS's hierarchical representation learning. Given an input $F_{i,0}$, the block applies $L$ sequential Swin Transformer Layers as shown in Eq.~\ref{eq:5}. In each STL, multi-head self-attention is performed within local windows of size $M\times M$. For a window feature $X \in \mathbb{R}^{M^2 \times C}$, query, key, and value matrices are computed as $Q = XW_Q,\quad K = XW_K,\quad V = XW_V$. Local attention is then evaluated as, $
\mathrm{Attention}(Q,K,V) = \mathrm{Softmax}\!\left(\frac{QK^\top}{\sqrt{d}} + B\right)V,$ where $B$ adds learnable relative positional encoding, this formulation enables the model to detect correlations between pixels that belong to the same semantic region (e.g., corners of the eyes or boundary of the mouth).

Window shifting significantly enhances the module's capacity by enabling cross-window interaction. Without this mechanism, signals would remain trapped within fixed windows, preventing the learning of long-range facial relationships. The alternating partition design ensures that information flows smoothly across the entire face. After the $L$ STLs, a convolution fuses the refined features before residual addition:
\begin{equation}
    F_{i,\mathrm{out}} = H_{\mathrm{Conv}_i}(F_{i,L}) + F_{i,0}.
\label{eq:7}
\end{equation}

This design offers two significant advantages.
\textit{Translational consistency}: The convolutional layer introduces spatially-invariant filtering, which complements the spatially-varying transformer attention. \textit{Identity preservation:} The residual skip ensures that essential structural information propagates across blocks without degradation. Thus, the RSTB forms a powerful hierarchical refinement module that models both fine structural details and global relationships.

\subsection{Reconstruction and Upsampling}

The refined feature map $F_{\mathrm{res}}$ is compressed with a channel-reduction convolution $F_{\mathrm{red}} = H_{\mathrm{red}}(F_{\mathrm{res}})$. Similarly, to increase the feature resolution, SwinIFS uses PixelShuffle, a highly efficient sub-pixel convolution that rearranges channels into spatial dimensions: $F_{\mathrm{up}} = \mathrm{PixelShuffle}_{S}\left(H_{\mathrm{up}}(F_{\mathrm{red}})\right)$. This operation produces a smooth, high-resolution feature map free from checkerboard artifacts common with transposed convolutions. A final $3\times3$ convolution maps the features to the RGB image domain, $\tilde{I}_{\mathrm{HR}} = H_{\mathrm{rec}}(F_{\mathrm{up}})$. To strengthen identity consistency and stabilize reconstruction, a global skip connection adds the bicubically upsampled LR image, $I_{\mathrm{HR}}^{\mathrm{final}} = \tilde{I}_{\mathrm{HR}} + \mathrm{Up}_{\mathrm{bicubic}}(I_{\mathrm{LR}})$. This ensures the preservation of global structure while the learned features restore missing high-frequency details.

\subsection{Loss Functions}

Training SwinIFS requires balancing pixel-level accuracy with perceptual realism. The primary objective is the $\ell_1$ reconstruction loss:
\begin{equation}
   \ell_1 = \frac{1}{N}\sum_{i=1}^{N} \|I_{\mathrm{HR}}^{(i)} - I_{\mathrm{GT}}^{(i)}\|_1,
   \label{eq:l1}
\end{equation}
which penalizes pixel-level deviations and encourages sharp, clean results. However, pixel-level losses do not fully capture perceptual similarity or the structure of identity. To address this, we incorporate a perceptual loss based on VGG-19 activations $\Phi$:

\begin{equation}
\ell_{\mathrm{VGG}} = \|\Phi(I_{\mathrm{HR}}) - \Phi(I_{\mathrm{GT}})\|_2^2.
   \label{eq:lvvg}
\end{equation}

This encourages the SR output to preserve semantic details such as eye shape, mouth curvature, skin texture, and other identity-related cues. The total loss driving model optimization is:
\begin{equation}
\ell_{\mathrm{total}} = \lambda_1 \ell_1 + \lambda_2 \ell_{\mathrm{VGG}}.
   \label{eq:total-loss}
\end{equation}

This hybrid objective ensures the network achieves both quantitative accuracy and perceptual quality, yielding reconstructed faces that are structurally consistent and visually realistic.
 
\section{Experiment}
\label{sec:experiement}
This section presents a comprehensive experimental evaluation of the proposed SwinIFS framework. The objective of the experiments is to assess reconstruction fidelity, perceptual realism, and structural consistency under challenging $4\times$ and $8\times$ upscaling scenarios. All experiments were designed to provide a fair and rigorous comparison with existing face super-resolution methods, accounting for both quantitative performance and visual quality. The evaluation protocol also aims to demonstrate the contribution of landmark-guided structural priors and hierarchical Swin Transformer modeling to identity preservation and fine-detail restoration.

\subsection{Dataset and Preprocessing}

All experiments were conducted on the CelebA dataset~\cite{liu2015celeba}, a widely used large-scale facial benchmark containing over 200,000 images of more than 10,000 identities. CelebA provides substantial diversity in pose, illumination, age, and facial attributes, along with five key facial landmarks, making it particularly suitable for landmark-guided face super-resolution. Each image is first aligned and then processed using a structure-aware cropping strategy inspired by DIC-Net~\cite{Ma_2020_CVPR}. A bounding box enclosing the five facial landmarks is expanded by a fixed margin to retain contextual facial regions, such as the hairline and jawline. The cropped images are resized to $128\times128$ to form the high-resolution supervision set.

Low-resolution images are synthesized via bicubic downsampling with scale factors $S=\{4,8\}$, yielding inputs of size $32\times32$ and $16\times16$, respectively. To incorporate structural priors, the five landmark coordinates for each LR image are converted into Gaussian heatmaps, thereby providing spatially continuous geometric guidance. These heatmaps are stacked with the LR RGB channels to form an eight-channel tensor that serves as the model input. A total of 168,854 images from the CelebA training split are used for model training, while 1,000 identity-disjoint images from the official test split are reserved for evaluation. This strict separation ensures unbiased generalization performance.

\subsection{Evaluation Metrics}

The evaluation of SwinIFS employs three widely accepted full-reference metrics: Peak Signal-to-Noise Ratio (PSNR), Structural Similarity Index Measure (SSIM), and Learned Perceptual Image Patch Similarity (LPIPS). PSNR quantifies pixel-level fidelity as the mean squared error between the reconstructed and ground-truth images. While higher PSNR generally reflects greater fidelity, it tends to correlate poorly with human perception, especially in tasks involving facial details and texture.

To complement PSNR, SSIM measures perceptual similarity between two images across luminance, contrast, and structural information. SSIM is computed on the luminance (Y) channel of the YCbCr color space, as luminance is most sensitive to visual distortions. LPIPS further extends perceptual assessment by comparing deep feature representations obtained from pretrained neural networks. This metric has been shown to correlate strongly with human judgments of perceptual similarity, making it particularly relevant for facial image restoration, where texture realism and identity consistency are crucial. Together, these metrics provide a balanced assessment of pixel-level accuracy, structural coherence, and perceptual quality.

\begin{table}[t]
\caption{
Quantitative comparison on the CelebA dataset for $4\times$ and $8\times$ 
face super-resolution. Best results are shown in \textcolor{red}{\textbf{bold}}, and 
second-best results are \textcolor{blue}{\underline{underlined}}.
}
\label{tab:quant_results}
\centering

\begin{tabular}{l|ccc|ccc}
\toprule
& \multicolumn{3}{c|}{$4\times$ Upscaling}
& \multicolumn{3}{c}{$8\times$ Upscaling} \\
\cmidrule(lr){2-4}
\cmidrule(lr){5-7}

\textbf{Model} & PSNR $\uparrow$ & SSIM $\uparrow$ & LPIPS & PSNR $\uparrow$ & SSIM $\uparrow$ & LPIPS \\
\midrule

Bicubic & 27.38 & 0.8002 & 0.1857 & 23.46 & 0.6776 & 0.2699 \\

SRGAN~\cite{ledig2017photo} 
& 31.05 & 0.8880 & 0.0459 
& 26.63 & 0.7628 & 0.1043 \\

FSRNet~\cite{chen2018fsrnet} 
& 31.37 & 0.9012 & 0.0501 
& 26.86 & 0.7714 & 0.1098 \\

DIC~\cite{Ma_2020_CVPR} 
& 31.58 & 0.9015 & 0.0532 
& 27.35 & \textcolor{blue}{\underline{0.8109}} & 0.0902 \\

SPARNet~\cite{sparnet} 
& 31.52 & 0.9005 & 0.0593 
& 27.29 & 0.7965 & 0.1088 \\

SISN~\cite{lu2021face} 
& 31.55 & 0.9010 & 0.0587 
& 26.83 & 0.7786 & 0.1044 \\

MRRNet~\cite{huang2022mrrnet} 
& 30.48 & 0.8720 & \textcolor{red}{\textbf{0.0374}} 
& 25.94 & 0.7417 & \textcolor{red}{\textbf{0.0562}} \\

WIPA~\cite{dastmalchi2022super} 
& 30.35 & 0.8711 & 0.0619 
& 26.23 & 0.7652 & 0.0961 \\

UFSRNet~\cite{wang2024ufsrnet} 
& 31.42 & 0.8987 & 0.0643 
& 27.10 & 0.7887 & 0.1102 \\

W-Net~\cite{liu2025w} 
& \textcolor{blue}{\underline{31.63}} & \textcolor{blue}{\underline{0.9029}} & 0.0425
& \textcolor{blue}{\underline{27.40}} & 0.8014 & 0.0760 \\

\textbf{SwinIFS (Ours)}
& \textcolor{red}{\textbf{32.01}} & \textcolor{red}{\textbf{0.9520}} & \textcolor{blue}{\underline{0.0404}}
& \textcolor{red}{\textbf{27.97}} & \textcolor{red}{\textbf{0.8513}} & \textcolor{blue}{\underline{0.0720}} \\

\bottomrule
\end{tabular}

\end{table}

\subsection{Implementation Details}

All experiments were conducted using PyTorch on a workstation equipped with dual NVIDIA RTX A6000 GPUs (48GB VRAM each). Mixed-precision FP16 computation was employed to reduce memory consumption and improve training efficiency. The SwinIFS model takes an eight-channel input comprising RGB image data and five landmark heatmaps. It processes the input through a shallow convolutional feature extractor, six Residual Swin Transformer Blocks, and a PixelShuffle-based reconstruction module to generate outputs at a resolution of $128\times128$.

The model is trained from scratch without any pretrained face SR weights. Convolutional layers are initialized using the default initialization, and transformer layers are initialized using truncated normal initialization to ensure stable optimization. Training is performed using the Adam optimizer with $(\beta_1,\beta_2)=(0.9,0.999)$ and an initial learning rate of $10^{-4}$. A MultiStepLR schedule is applied with decay milestones at 250,000 and 400,000 iterations.The loss function is a weighted combination of $\ell_1$ reconstruction loss and VGG-based perceptual loss, with weights $(\lambda_{\ell_1},\lambda_2)=(1.0,0.1)$.

All images are normalized to the $[0,1]$ range, and no data augmentation techniques, such as flipping or rotation, are applied. To maintain deterministic behavior, NumPy, PyTorch, and CUDA are configured with fixed random seeds, and CuDNN is configured in deterministic mode. Periodic checkpointing and TensorBoard logging are used to monitor loss curves and evaluation metrics throughout training. This experimental configuration ensures reproducibility, fairness, and consistency across all comparisons presented in the following sections. The results reported next highlight the strengths of SwinIFS in both objective and perceptual evaluation settings.

\begin{figure}[t]
\centering
\begin{tabular}{cccccc}
\includegraphics[width=0.15\textwidth]{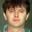}&
\includegraphics[width=0.15\textwidth]{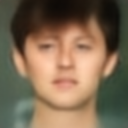}&
\includegraphics[width=0.15\textwidth]{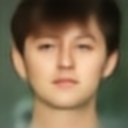}& 
\includegraphics[width=0.15\textwidth]{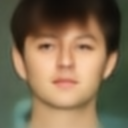}& 
\includegraphics[width=0.15\textwidth]{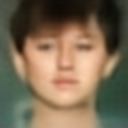}&
\includegraphics[width=0.15\textwidth]{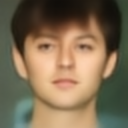}  \\
LR   & SRGAN &FSRNET & DIC & SPARNET & SISN\\ 

\includegraphics[width=0.15\textwidth]{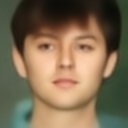}&
\includegraphics[width=0.15\textwidth]{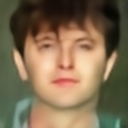}&
\includegraphics[width=0.15\textwidth]{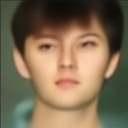}&
\includegraphics[width=0.15\textwidth]{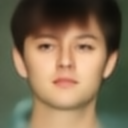}&
\includegraphics[width=0.15\textwidth]{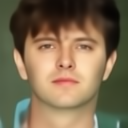}&
\includegraphics[width=0.15\textwidth]{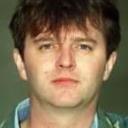}     \\
MRRNET   & WIPA &UFSRNET & W-NET & SWINIFS & HR \\ 
\end{tabular}
\caption{Visual comparison of face super-resolution results for $4\times$ upscaling on CelebA. SwinIFS produces sharper and more identity-preserving reconstructions than competing methods.}
\label{fig:x4_visual_1}
\end{figure}

\section{Results and Discussion}
The performance of the proposed SwinIFS framework is assessed through extensive quantitative and qualitative comparisons against a wide range of state-of-the-art face super-resolution methods. These include classical CNN-based models such as SRCNN and FSRNet, GAN-based methods such as SRGAN and SPARNet, and more recent landmark-aware and Transformer-based methods, including DIC, WIPA, UFSRNet, and W-Net. All evaluations are performed under identical conditions using the CelebA test set, ensuring a fair and consistent benchmarking environment.

\vspace{1mm}\noindent\textbf{Quantitative Comparisons:} The quantitative results presented in Table~\ref{tab:quant_results} demonstrate that SwinIFS achieves the highest PSNR and SSIM values for both $4\times$ and $8\times$ upscaling, while maintaining one of the lowest LPIPS scores. These results indicate that SwinIFS excels at both pixel-level fidelity and perceptual similarity, which are essential for restoring realistic, identity-consistent facial details. The improvements are particularly pronounced at $8\times$ upscaling, where most methods struggle due to severe information loss. SwinIFS achieves a PSNR of 27.97dB and an SSIM of 0.851, outperforming recent Transformer-based competitors such as W-Net and UFSRNet, and significantly surpassing classical CNN or GAN-based approaches. The low LPIPS score further underscores SwinIFS's perceptual advantage, reflecting its ability to generate natural textures without introducing GAN-related artifacts.

\begin{figure}[t]
\centering
\begin{tabular}{cccccc}
\includegraphics[width=0.15\textwidth]{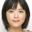}&
\includegraphics[width=0.15\textwidth]{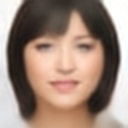}&
\includegraphics[width=0.15\textwidth]{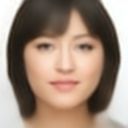}& 
\includegraphics[width=0.15\textwidth]{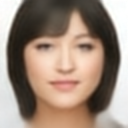}& 
\includegraphics[width=0.15\textwidth]{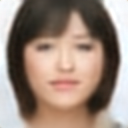}&
\includegraphics[width=0.15\textwidth]{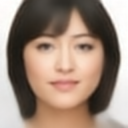}  \\

&
\includegraphics[trim={60pt 70pt 30pt 20pt}, clip, width=0.15\textwidth]{Figs/Fig2/X4-202409/202409_SRGAN.png}&
\includegraphics[trim={60pt 70pt 30pt 20pt}, clip, width=0.15\textwidth]{Figs/Fig2/X4-202409/202409_FSRNET.png}& 
\includegraphics[trim={60pt 70pt 30pt 20pt}, clip, width=0.15\textwidth]{Figs/Fig2/X4-202409/202409_DIC.png}& 
\includegraphics[trim={60pt 70pt 30pt 20pt}, clip, width=0.15\textwidth]{Figs/Fig2/X4-202409/202409_SPARNET.png}&
\includegraphics[trim={60pt 70pt 30pt 20pt}, clip, width=0.15\textwidth]{Figs/Fig2/X4-202409/202409_SISN.png}  \\
LR   & SRGAN &FSRNET & DIC & SPARNET & SISN\\ 

\includegraphics[width=0.15\textwidth]{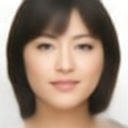}&
\includegraphics[width=0.15\textwidth]{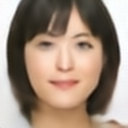}&
\includegraphics[width=0.15\textwidth]{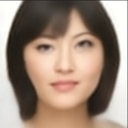}&
\includegraphics[width=0.15\textwidth]{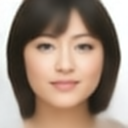}&
\includegraphics[width=0.15\textwidth]{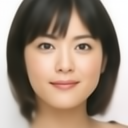}&
\includegraphics[width=0.15\textwidth]{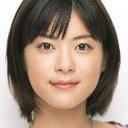}     \\

\includegraphics[trim={60pt 70pt 30pt 20pt}, clip, width=0.15\textwidth]{Figs/Fig2/X4-202409/202409_MRRNET.png}&
\includegraphics[trim={60pt 70pt 30pt 20pt}, clip, width=0.15\textwidth]{Figs/Fig2/X4-202409/202409_WIPA.png}&
\includegraphics[trim={60pt 70pt 30pt 20pt}, clip, width=0.15\textwidth]{Figs/Fig2/X4-202409/202409_UFSRNET.png}&
\includegraphics[trim={60pt 70pt 30pt 20pt}, clip, width=0.15\textwidth]{Figs/Fig2/X4-202409/202409_W-NET.png}&
\includegraphics[trim={60pt 70pt 30pt 20pt}, clip, width=0.15\textwidth]{Figs/Fig2/X4-202409/202409_SWINIFS.png}&
\includegraphics[trim={60pt 70pt 30pt 20pt}, clip, width=0.15\textwidth]{Figs/Fig2/X4-202409/202409-HR.jpg}     \\
MRRNET   & WIPA &UFSRNET & W-NET & SWINIFS & HR \\ 
\end{tabular}
\caption{A qualitative comparison of four times face super-resolution results on CelebA. Compared with other state-of-the-art methods, SwinIFS produces reconstructions that are both visually sharper and more effective in maintaining facial identity. The eye is clear, with no deformation or loss of detail, unlike competing state-of-the-art methods.}
\label{fig:x4_visual_2}
\end{figure}

\begin{figure}[t]
\centering
\begin{tabular}{cccccc}
\includegraphics[width=0.15\textwidth]{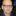}&
\includegraphics[width=0.15\textwidth]{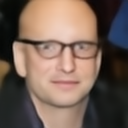}&
\includegraphics[width=0.15\textwidth]{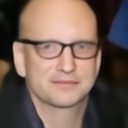}& 
\includegraphics[width=0.15\textwidth]{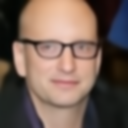}& 
\includegraphics[width=0.15\textwidth]{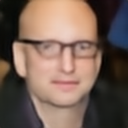}&
\includegraphics[width=0.15\textwidth]{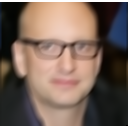}  \\
LR   & SRGAN &FSRNET & DIC & SPARNET & SISN\\ 

\includegraphics[width=0.15\textwidth]{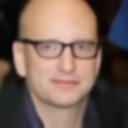}&
\includegraphics[width=0.15\textwidth]{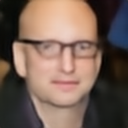}&
\includegraphics[width=0.15\textwidth]{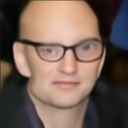}&
\includegraphics[width=0.15\textwidth]{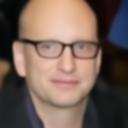}&
\includegraphics[width=0.15\textwidth]{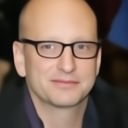}&
\includegraphics[width=0.15\textwidth]{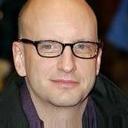}     \\
MRRNET   & WIPA &UFSRNET & W-NET & SWINIFS & HR \\ 
\end{tabular}
\caption{Visual comparison of face super-resolution results for 8$\times$ upscaling on CelebA.  Most baselines exhibit blurred contours and a loss of identity-defining structure, whereas our model recovers coherent global facial geometry, the contour of the glasses, and sharper details around the eyes, nose, and lips.}
\label{fig:x8_visual_1}
\end{figure}

\begin{figure}[t]
\centering
\begin{tabular}{cccccc}
\includegraphics[width=0.15\textwidth]{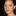}&
\includegraphics[width=0.15\textwidth]{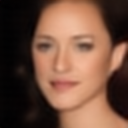}&
\includegraphics[width=0.15\textwidth]{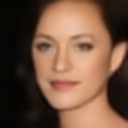}& 
\includegraphics[width=0.15\textwidth]{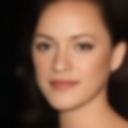}& 
\includegraphics[width=0.15\textwidth]{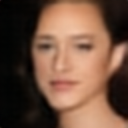}&
\includegraphics[width=0.15\textwidth]{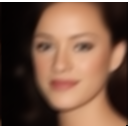}  \\

&
\includegraphics[trim={40pt 30pt 30pt 70pt}, clip, width=0.15\textwidth]{Figs/Fig3/X8-200788/200788_SRGAN.png}&
\includegraphics[trim={40pt 30pt 30pt 70pt}, clip, width=0.15\textwidth]{Figs/Fig3/X8-200788/200788_FSRNET.png}& 
\includegraphics[trim={40pt 30pt 30pt 70pt}, clip, width=0.15\textwidth]{Figs/Fig3/X8-200788/200788-DIC.png}& 
\includegraphics[trim={40pt 30pt 30pt 70pt}, clip, width=0.15\textwidth]{Figs/Fig3/X8-200788/200788_SPARNET.png}&
\includegraphics[trim={40pt 30pt 30pt 70pt}, clip, width=0.15\textwidth]{Figs/Fig3/X8-200788/200788-SISN.png}  \\
LR   & SRGAN &FSRNET & DIC & SPARNET & SISN\\ 

\includegraphics[width=0.15\textwidth]{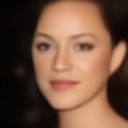}&
\includegraphics[width=0.15\textwidth]{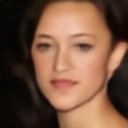}&
\includegraphics[width=0.15\textwidth]{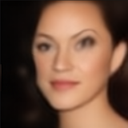}&
\includegraphics[width=0.15\textwidth]{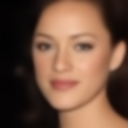}&
\includegraphics[width=0.15\textwidth]{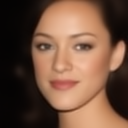}&
\includegraphics[width=0.15\textwidth]{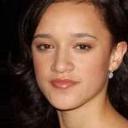}     \\

\includegraphics[trim={40pt 30pt 30pt 70pt}, clip, width=0.15\textwidth]{Figs/Fig3/X8-200788/200788_MRRNET.png}&
\includegraphics[trim={40pt 30pt 30pt 70pt}, clip, width=0.15\textwidth]{Figs/Fig3/X8-200788/200788_WIPA.png}&
\includegraphics[trim={40pt 30pt 30pt 70pt}, clip, width=0.15\textwidth]{Figs/Fig3/X8-200788/200788_UFSRNET.png}&
\includegraphics[trim={40pt 30pt 30pt 70pt}, clip, width=0.15\textwidth]{Figs/Fig3/X8-200788/200788-W-NET.png}&
\includegraphics[trim={40pt 30pt 30pt 70pt}, clip, width=0.15\textwidth]{Figs/Fig3/X8-200788/200788_SWINIFS.png}&
\includegraphics[trim={40pt 30pt 30pt 70pt}, clip, width=0.15\textwidth]{Figs/Fig3/X8-200788/200788-HR.jpg}     \\
MRRNET   & WIPA &UFSRNET & W-NET & SWINIFS & HR \\ 
\end{tabular}
\caption{Regional comparison of mouth reconstruction for 8$\times$ upscaling on CelebA. Competing approaches tend to generate smudged or excessively smooth lip textures, whereas SwinIFS maintains the natural shading, lip curvature, and texture continuity, thereby aligning more accurately with the ground truth.}
\label{fig:x8_visual_2}
\end{figure}

\vspace{1mm}\noindent\textbf{Qualitative Comparisons at 4$\times$ Scale:} The qualitative comparisons in Figure~\ref{fig:x4_visual_1} and~\ref{fig:x4_visual_2} further substantiate these findings. For $4\times$ upscaling, SwinIFS reconstructs faces with sharper contours, clearer eye regions, and more realistic mouth textures than competing models. Many CNN- and GAN-based baselines produce overly smooth or plastic-like textures, whereas recent architectures often hallucinate details that distort identity. In contrast, SwinIFS restores features in a manner that remains faithful to the ground truth, owing to its integration of landmark-guided geometric priors and hierarchical attention mechanisms.

To further examine the reconstruction of identity-critical regions, Figure~\ref{fig:x4_visual_2} presents region-specific comparisons focusing on the eyes. These areas are particularly challenging because they contain fine structural cues essential for identity recognition. SwinIFS restores sharper eye boundaries, more accurate iris structure, and more realistic eyelid geometry than all other baselines.


\vspace{1mm}\noindent\textbf{Qualitative Comparisons at 8$\times$ Scale:} At the more challenging $8\times$ scale, illustrated in Figure~\ref{fig:x8_visual_1}, most competing models fail to recover meaningful structure from the highly degraded LR inputs. Their outputs exhibit blurred contours, incorrect shape reconstruction, and a loss of characteristic identity cues. SwinIFS, by contrast, reconstructs the global facial geometry while restoring high-frequency detail around the eyes, nose, and lips. The advantage is most evident in cases involving significant pose variations or harsh illumination, where our model preserves identity coherence and reduces artifacts. Similarly, in Figure~\ref{fig:x8_visual_2}, the mouth region reconstructed by SwinIFS retains natural shading, lip curvature, and texture continuity, avoiding the smudging or over-smoothing seen in alternative approaches. These region-focused comparisons confirm that incorporating landmark heatmaps enables SwinIFS to allocate attention effectively to semantically meaningful facial regions, thereby improving structural fidelity.

\begin{figure}[t]
\centering
\includegraphics[width=0.65\linewidth]{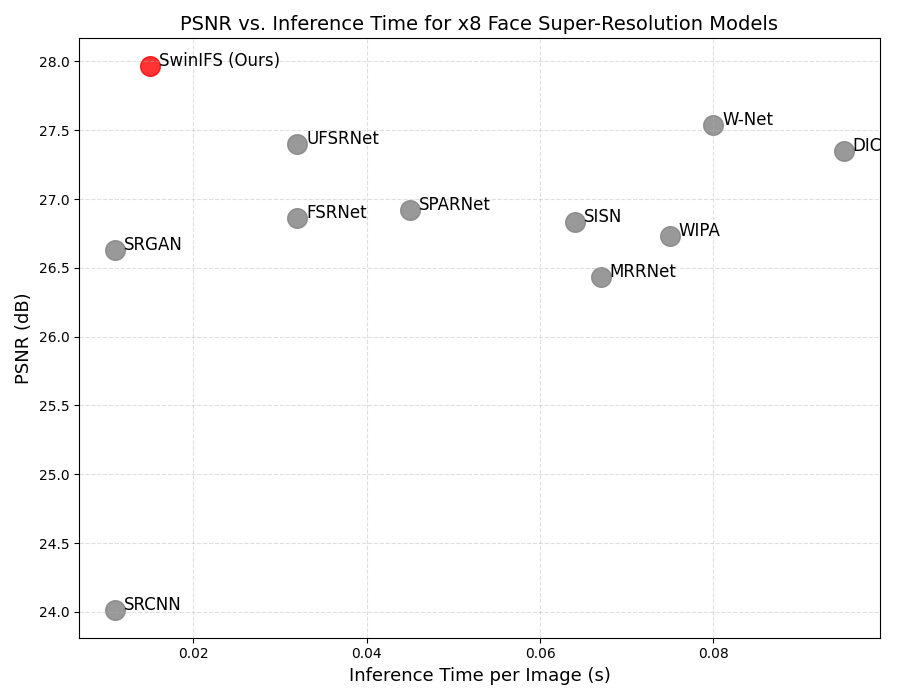}
\caption{PSNR versus inference time for $8\times$ face super-resolution models. SwinIFS achieves the best balance between reconstruction quality and computational efficiency.}
\label{fig:psnr_time_plot}
\end{figure}

Beyond reconstruction quality, practical face super-resolution systems must also consider computational efficiency. Figure~\ref{fig:psnr_time_plot} presents the relationship between PSNR and inference time for a range of competing models. While methods such as DIC, WIPA, and W-Net achieve competitive PSNR values, they incur significantly higher inference times due to deeper architectures or multi-stage refinement. Lightweight models such as SRCNN and SRGAN offer faster inference but fail to recover detailed and identity-relevant structures, especially at higher upscaling factors. SwinIFS achieves a favorable balance between accuracy and efficiency, reaching the highest PSNR while maintaining an inference time of only 0.015 seconds per $128\times128$ face. This positions SwinIFS on the Pareto frontier, demonstrating that its design, built on efficient window-based attention and compact hierarchical blocks, supports both real-time performance and high-fidelity reconstruction.

Taken together, the results clearly show that SwinIFS achieves a superior combination of perceptual realism, structural consistency, and computational efficiency compared to prior state-of-the-art methods. The model’s ability to integrate geometric priors, hierarchical feature refinement, and efficient Transformer-based modeling enables it to produce photorealistic, identity-faithful results even at extreme magnification. The consistency of improvements across quantitative metrics, global visual comparisons, and region-specific analyses demonstrates the robustness and reliability of SwinIFS for real-world face super-resolution applications.

\section{Conclusion}

This work introduces SwinIFS, a landmark-guided face super-resolution framework that addresses the challenges posed by severely degraded facial inputs. By integrating dense structural priors with a hierarchical Swin Transformer backbone, the proposed method effectively recovers fine‐grained textures while preserving global facial geometry and identity. Extensive experiments on the CelebA dataset demonstrate that SwinIFS consistently outperforms existing CNN, GAN, and Transformer-based approaches across both $4\times$ and $8\times$ upscaling factors. The model achieves superior quantitative performance and produces visually convincing high-resolution reconstructions, particularly in identity-critical regions such as the eyes and mouth. Moreover, SwinIFS offers a favorable trade-off between accuracy and inference speed, making it suitable for real-world applications where both quality and efficiency are essential. 

\vspace{1mm}\noindent\textbf{Limitation:}
While the framework demonstrates strong robustness, it still relies on accurate landmark predictions and has been evaluated primarily on frontal and near-frontal facial images. While SwinIFS demonstrates strong performance across upscaling factors, several limitations remain. Our framework depends on accurate facial landmark detection; under severe blur, occlusion, or non-frontal poses, landmark localization becomes unreliable, and errors propagate directly into the Gaussian heatmap priors, potentially misguiding the attention mechanism. Similarly, our and the competing method's evaluation is restricted to CelebA, which is dominated by frontal and near-frontal, well-illuminated faces; performance under large pose variation, heavy occlusion (e.g., masks, sunglasses), or real-world sensor degradation (as opposed to synthetic bicubic downsampling) remains untested. Lastly, the five-point landmark set used here captures only coarse facial geometry; finer structures such as wrinkles, facial hair, or asymmetric expressions are not explicitly modeled. 

\vspace{1mm}\noindent\textbf{Future Direction:} The above-mentioned limitation could be jointly and iteratively refined by refining landmark detection and super-resolution, extending the five-point landmark scheme to denser or parsing-based structural priors, and training on more diverse, real-world degraded datasets spanning varied poses and occlusions. Additionally, evaluating the model on multi-domain or real-world degraded datasets would strengthen its applicability. Overall, SwinIFS presents a significant step toward reliable, identity-preserving face super-resolution and provides a strong foundation for further advancements in facial enhancement technologies.

%
%
\bibliographystyle{splncs04}
\bibliography{main-accv}

\begin{center}
    {\bf Supplementary Materials}
\end{center}

\section{Ablations}
\begin{figure}[t]
\centering
\begin{tabular}{ccccc}
\includegraphics[width=0.15\textwidth]{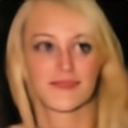}&
\includegraphics[width=0.15\textwidth]{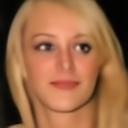}&
\includegraphics[width=0.15\textwidth]{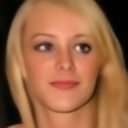}&
\includegraphics[width=0.15\textwidth]{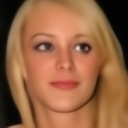}&
\includegraphics[width=0.15\textwidth]{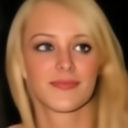}  \\

\includegraphics[width=0.15\textwidth]{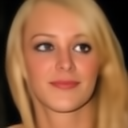}&
\includegraphics[width=0.15\textwidth]{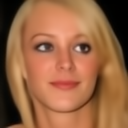}&
\includegraphics[width=0.15\textwidth]{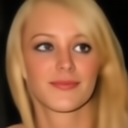}&
\includegraphics[width=0.15\textwidth]{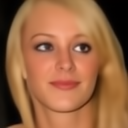}&
\includegraphics[width=0.15\textwidth]{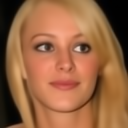}  \\

\includegraphics[width=0.15\textwidth]{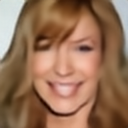}&
\includegraphics[width=0.15\textwidth]{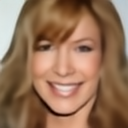}&
\includegraphics[width=0.15\textwidth]{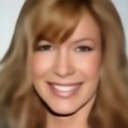}&
\includegraphics[width=0.15\textwidth]{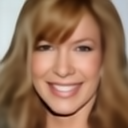}&
\includegraphics[width=0.15\textwidth]{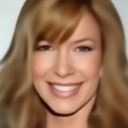}  \\

\includegraphics[width=0.15\textwidth]{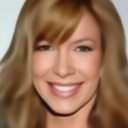}&
\includegraphics[width=0.15\textwidth]{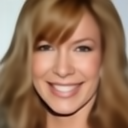}&
\includegraphics[width=0.15\textwidth]{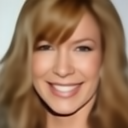}&
\includegraphics[width=0.15\textwidth]{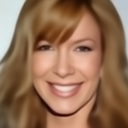}&
\includegraphics[width=0.15\textwidth]{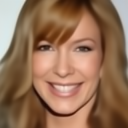}  \\

\includegraphics[width=0.15\textwidth]{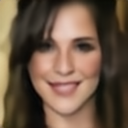}&
\includegraphics[width=0.15\textwidth]{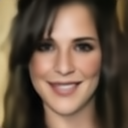}&
\includegraphics[width=0.15\textwidth]{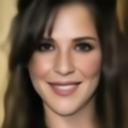}&
\includegraphics[width=0.15\textwidth]{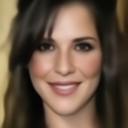}&
\includegraphics[width=0.15\textwidth]{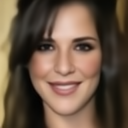}  \\

\includegraphics[width=0.15\textwidth]{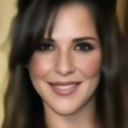}&
\includegraphics[width=0.15\textwidth]{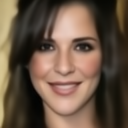}&
\includegraphics[width=0.15\textwidth]{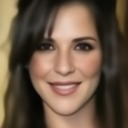}&
\includegraphics[width=0.15\textwidth]{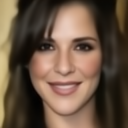}&
\includegraphics[width=0.15\textwidth]{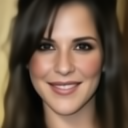}  \\
\end{tabular}
\caption{$4\times$ shows the effectiveness of the landmark-guided framework.}
\label{fig:4x_ablation}
\end{figure}

\begin{figure}[t]
\centering
\begin{tabular}{ccccc}
\includegraphics[width=0.15\textwidth]{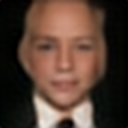}&
\includegraphics[width=0.15\textwidth]{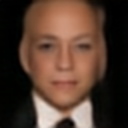}&
\includegraphics[width=0.15\textwidth]{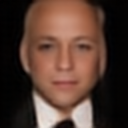}&
\includegraphics[width=0.15\textwidth]{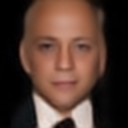}&
\includegraphics[width=0.15\textwidth]{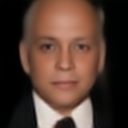}  \\

\includegraphics[width=0.15\textwidth]{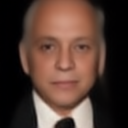}&
\includegraphics[width=0.15\textwidth]{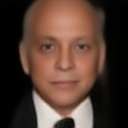}&
\includegraphics[width=0.15\textwidth]{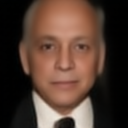}&
\includegraphics[width=0.15\textwidth]{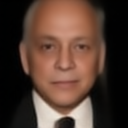}&
\includegraphics[width=0.15\textwidth]{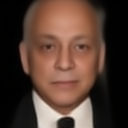}  \\

\includegraphics[width=0.15\textwidth]{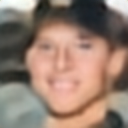}&
\includegraphics[width=0.15\textwidth]{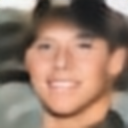}&
\includegraphics[width=0.15\textwidth]{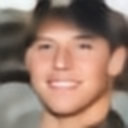}&
\includegraphics[width=0.15\textwidth]{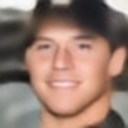}&
\includegraphics[width=0.15\textwidth]{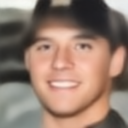}  \\

\includegraphics[width=0.15\textwidth]{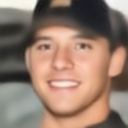}&
\includegraphics[width=0.15\textwidth]{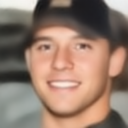}&
\includegraphics[width=0.15\textwidth]{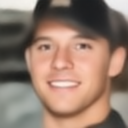}&
\includegraphics[width=0.15\textwidth]{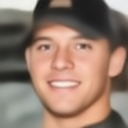}&
\includegraphics[width=0.15\textwidth]{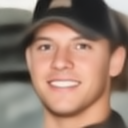}  \\

\includegraphics[width=0.15\textwidth]{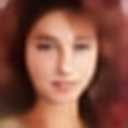}&
\includegraphics[width=0.15\textwidth]{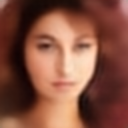}&
\includegraphics[width=0.15\textwidth]{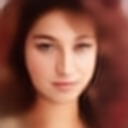}&
\includegraphics[width=0.15\textwidth]{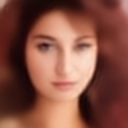}&
\includegraphics[width=0.15\textwidth]{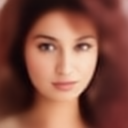}  \\

\includegraphics[width=0.15\textwidth]{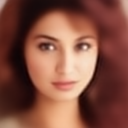}&
\includegraphics[width=0.15\textwidth]{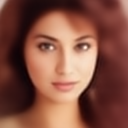}&
\includegraphics[width=0.15\textwidth]{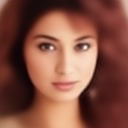}&
\includegraphics[width=0.15\textwidth]{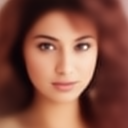}&
\includegraphics[width=0.15\textwidth]{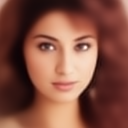}  \\
\end{tabular}
\caption{$8\times$ shows the effectiveness of the landmark-guided framework.}
\label{fig:8x_ablation}
\end{figure}

To further analyze the effectiveness of the proposed landmark-guided framework, we conduct a qualitative ablation study in the challenging $4\times$ and $8\times$ face super-resolution settings. Figures~\ref{fig:4x_ablation} and \ref{fig:8x_ablation} present reconstruction results obtained from the same low-resolution input at different training stages. Although the landmark heatmaps are extracted only once and used as structural guidance throughout the reconstruction process, the progressive reconstruction results provide insight into how the network learns to exploit the landmark information during optimization.
As shown in Figures~\ref{fig:4x_ablation} and \ref{fig:8x_ablation}, the reconstructed facial images exhibit blurred boundaries and incomplete facial structures during the early stages of training. As the optimization proceeds, the network progressively restores identity-sensitive facial components, including the eyes, nose, mouth, and facial contours. The final reconstruction preserves sharper facial details and more coherent facial geometry, indicating that the structural information provided by the landmark heatmaps effectively guides the feature learning process throughout training. The qualitative improvements observed across the training stages suggest that the landmark-guided representation enables the proposed SwinIFS framework to better preserve facial structure and recover fine-grained details under severe $8\times$ degradation. These results demonstrate that incorporating facial structural priors facilitates the progressive refinement of facial features and improves reconstruction quality in the final model.

\section{Visualizations}
\begin{figure}[t]
\centering
\begin{tabular}{ccccc}
\includegraphics[width=0.15\textwidth]{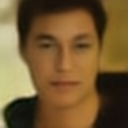}&
\includegraphics[width=0.15\textwidth]{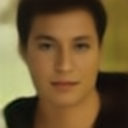}& 
\includegraphics[width=0.15\textwidth]{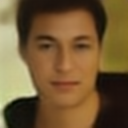}& 
\includegraphics[width=0.15\textwidth]{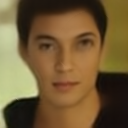}&
\includegraphics[width=0.15\textwidth]{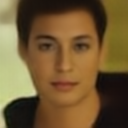}  \\

\includegraphics[trim={30pt 70pt 30pt 30pt}, clip, width=0.15\textwidth]{Suppl-Figs/vis-figs/X4-VIS/202566/202566_SRGAN.png}&
\includegraphics[trim={30pt 70pt 30pt 30pt}, clip, width=0.15\textwidth]{Suppl-Figs/vis-figs/X4-VIS/202566/202566_FSRNET.png}& 
\includegraphics[trim={30pt 70pt 30pt 30pt}, clip, width=0.15\textwidth]{Suppl-Figs/vis-figs/X4-VIS/202566/202566_DIC.png}& 
\includegraphics[trim={30pt 70pt 30pt 30pt}, clip, width=0.15\textwidth]{Suppl-Figs/vis-figs/X4-VIS/202566/202566_UFSRNET.png}&
\includegraphics[trim={30pt 70pt 30pt 30pt}, clip, width=0.15\textwidth]{Suppl-Figs/vis-figs/X4-VIS/202566/202566_SISN.png}  \\
SRGAN &FSRNET & DIC & UFSRNET & SISN\\ 

\includegraphics[width=0.15\textwidth]{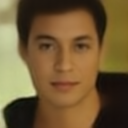}&
\includegraphics[width=0.15\textwidth]{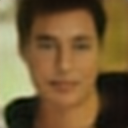}&
\includegraphics[width=0.15\textwidth]{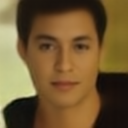}&
\includegraphics[width=0.15\textwidth]{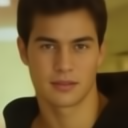}&
\includegraphics[width=0.15\textwidth]{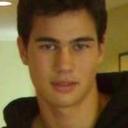}     \\

\includegraphics[trim={30pt 70pt 30pt 30pt}, clip, width=0.15\textwidth]{Suppl-Figs/vis-figs/X4-VIS/202566/202566_MRRNET.png}&
\includegraphics[trim={30pt 70pt 30pt 30pt}, clip, width=0.15\textwidth]{Suppl-Figs/vis-figs/X4-VIS/202566/202566_WIPA.png}&
\includegraphics[trim={30pt 70pt 30pt 30pt}, clip, width=0.15\textwidth]{Suppl-Figs/vis-figs/X4-VIS/202566/202566_W-NET.png}&
\includegraphics[trim={30pt 70pt 30pt 30pt}, clip, width=0.15\textwidth]{Suppl-Figs/vis-figs/X4-VIS/202566/202566_SWINIFS.png}&
\includegraphics[trim={30pt 70pt 30pt 30pt}, clip, width=0.15\textwidth]{Suppl-Figs/vis-figs/X4-VIS/202566/202566-HR.jpg}     \\
MRRNET   & WIPA &  W-NET & SWINIFS & HR \\ 
\end{tabular}
\caption{$4\times$ face super-resolution results on CelebA. In comparison to other leading methodologies, our approach provides reconstructions that are markedly sharper and superior in maintaining facial identity. The eye retains clarity, devoid of deformation or detail loss, in contrast to other techniques.}
\label{fig:x4_visual_11}
\end{figure}

\begin{figure}[t]
\centering
\begin{tabular}{ccccc}
\includegraphics[width=0.15\textwidth]{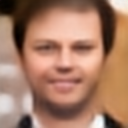}&
\includegraphics[width=0.15\textwidth]{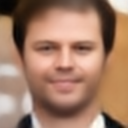}& 
\includegraphics[width=0.15\textwidth]{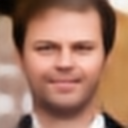}& 
\includegraphics[width=0.15\textwidth]{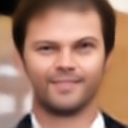}&
\includegraphics[width=0.15\textwidth]{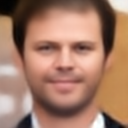}  \\

\includegraphics[trim={30pt 30pt 30pt 60pt}, clip, width=0.15\textwidth]{Suppl-Figs/vis-figs/X4-VIS/202581/202581_SRGAN.png}&
\includegraphics[trim={30pt 30pt 30pt 60pt}, clip, width=0.15\textwidth]{Suppl-Figs/vis-figs/X4-VIS/202581/202581_FSRNET.png}& 
\includegraphics[trim={30pt 30pt 30pt 60pt}, clip, width=0.15\textwidth]{Suppl-Figs/vis-figs/X4-VIS/202581/202581_DIC.png}& 
\includegraphics[trim={30pt 30pt 30pt 60pt}, clip, width=0.15\textwidth]{Suppl-Figs/vis-figs/X4-VIS/202581/202581_UFSRNET.png}&
\includegraphics[trim={30pt 30pt 30pt 60pt}, clip, width=0.15\textwidth]{Suppl-Figs/vis-figs/X4-VIS/202581/202581_SISN.png}  \\
SRGAN &FSRNET & DIC & UFSRNET & SISN\\ 

\includegraphics[width=0.15\textwidth]{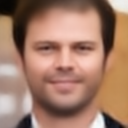}&
\includegraphics[width=0.15\textwidth]{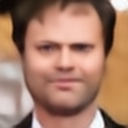}&
\includegraphics[width=0.15\textwidth]{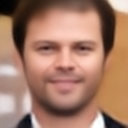}&
\includegraphics[width=0.15\textwidth]{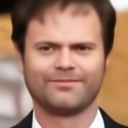}&
\includegraphics[width=0.15\textwidth]{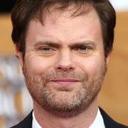}     \\

\includegraphics[trim={30pt 30pt 30pt 60pt}, clip, width=0.15\textwidth]{Suppl-Figs/vis-figs/X4-VIS/202581/202581_MRRNET.png}&
\includegraphics[trim={30pt 30pt 30pt 60pt}, clip, width=0.15\textwidth]{Suppl-Figs/vis-figs/X4-VIS/202581/202581_WIPA.png}&
\includegraphics[trim={30pt 30pt 30pt 60pt}, clip, width=0.15\textwidth]{Suppl-Figs/vis-figs/X4-VIS/202581/202581_W-NET.png}&
\includegraphics[trim={30pt 30pt 30pt 60pt}, clip, width=0.15\textwidth]{Suppl-Figs/vis-figs/X4-VIS/202581/202581_SWINIFS.png}&
\includegraphics[trim={30pt 30pt 30pt 60pt}, clip, width=0.15\textwidth]{Suppl-Figs/vis-figs/X4-VIS/202581/202581-HR.jpg}     \\
MRRNET   & WIPA &  W-NET & SWINIFS & HR \\ 
\end{tabular}
\caption{Compared to other advanced techniques, SwinIFS provides reconstructions that are visually sharper and more effective at maintaining facial identity. The eye appears clear, free from any deformation or loss of detail, in contrast to competing methodologies.}
\label{fig:x4_visual_22}
\end{figure}

We provide more visualization examples in Figure~\ref{fig:x4_visual_11} and \ref{fig:x4_visual_22}.

\end{document}